\documentclass{article}

\usepackage{microtype}
\usepackage{graphicx}
\usepackage{subfigure}
\usepackage{booktabs} 
\usepackage{multirow}
\usepackage{amsfonts,amssymb}
\usepackage{amsmath}
\usepackage{mathabx}
\usepackage{hyperref}

\usepackage{tabularx}
\usepackage{booktabs} 

\newcolumntype{M}[1]{>{\centering\arraybackslash}m{#1}} 
\newcolumntype{R}[1]{>{\raggedleft\arraybackslash}m{#1}} 
\newcolumntype{L}[1]{>{\raggedright\arraybackslash}m{#1}} 

\usepackage[accepted]{icml2021}

\icmltitlerunning{Switch Spaces}

\begin{document}

\twocolumn[
\icmltitle{Switch Spaces: Learning Product Spaces with Sparse Gating}



\icmlsetsymbol{equal}{*}

\begin{icmlauthorlist}
\icmlauthor{Shuai Zhang}{eth}
\icmlauthor{Yi Tay}{goo}
\icmlauthor{Wenqi Jiang}{eth}
\icmlauthor{Da-cheng Juan}{goo}
\icmlauthor{Ce Zhang}{eth}
\end{icmlauthorlist}

\icmlaffiliation{eth}{Department of Computer Science, ETH Zurich, Switzerland}
\icmlaffiliation{goo}{Google Research, United States}

\icmlcorrespondingauthor{Shuai Zhang}{shuai.zhang@inf.ethz.ch}

\icmlkeywords{Machine Learning, ICML}

\vskip 0.3in
]



\printAffiliationsAndNotice{}  

\begin{abstract}



Learning embedding spaces of suitable geometry is critical for representation learning. In order for learned representations to be effective and efficient, it is ideal that the geometric inductive bias aligns well with the underlying structure of the data. In this paper, we propose Switch Spaces, a data-driven approach for learning representations in product space. Specifically, product spaces (or manifolds) are spaces of mixed curvature, i.e., a combination of multiple euclidean and non-euclidean (hyperbolic, spherical) manifolds. To this end, we introduce sparse gating mechanisms that learn to choose, combine and switch spaces, allowing them to be switchable depending on the input data with specialization. Additionally, the proposed method is also efficient and has a constant computational complexity regardless of the model size. Experiments on knowledge graph completion and item recommendations show that the proposed switch space achieves new state-of-the-art performances, outperforming pure product spaces and recently proposed task-specific models. 


\end{abstract}

\section{Introduction}
\label{submission}
Many machine learning applications involve representation learning~\cite{bengio2013representation}. Euclidean and recently non-Euclidean spaces have demonstrated successful applications in modeling data with certain structures in various fields. Euclidean space has been extensively studied in the literature and has been serving as the workhorse for decades in representation learning~\cite{bordes2013translating,mikolov2013efficient,hsieh2017collaborative}. Non-Euclidean space has gained increasing attention as it excels at modeling data with hierarchical or cyclical structures~\cite{wilson2014spherical,nickel2017poincare,chami-etal-2020-low,vinh2020hyperml,meng2019spherical,ganea2018hyperbolic}.  Until recently, combining multiple spaces (component spaces) to learn product representations has been proposed and has shown promising results~\cite{gu2018learning,skopek2019mixed}. The main reason for this design is that our data is generated with multiple structures and has varying patterns, rather than be uniformly structured. Models with single type of manifold usually fail in capturing the underlying structures and can cause higher distortion in representation.  Product space is an ideal choice as it mixes different types of geometric models as an ensemble which is more suitable for complex and intricate input data.



However, in the existing product-space models, after the spaces are determined (e.g., spherical space $\bigtimes$ hyperbolic space), training samples are used to fit the learnable parameters in all these spaces, regardless of the appropriateness (e.g., both spherical and hyperbolic spaces will be updated even if the structure of inputs only aligns with the spherical space). It is difficult to align the geometry of input data with its suitable manifolds via simple combination of spaces without any specialization. Clearly, samples with different geometric structures should be handled with different component spaces. It is essential for component space models to calibrate their internal parameters by focusing on the subset of the training cases that fit the manifold and ignore cases they are not good at modeling. Nevertheless, the intricacy of real-world data makes it impossible to explicitly obtain the exact underlying geometry, let alone partition them into subsets.

To tackle this problem, we propose switch spaces, a data-driven representation approach that automatically chooses suitable spaces for each input data point based on the input-output relationship. Specifically, a sparse gating mechanism is introduced for automatic spaces selection. In this framework, only a certain number of spaces are active in calculation for each incoming sample, which encourages specialization and modularity. Moreover, it enables exploration of various combinations of spaces. Concretely, it can construct a number of smaller product spaces with different signatures based on the given data, allowing to display a greater variety of behaviors. We redefine the scoring function in switch spaces which subsumes the squared distances of product spaces but offers more flexibility.  In addition, the computation cost of switch spaces is linear to the number of active spaces, thus making building larger models possible. 






The contributions of this paper are: (a) We propose a representation learning framework, switch spaces, which learn to choose, combine and switch spaces via a sparse gating mechanism. It effectively aligns different types of geometric spaces with the characteristics of the input data, while being computationally efficient; (b) We construct models with switch spaces for two tasks including knowledge graph completion and item recommendation and verify its effectiveness with four real-world datasets. Empirical experiments show that switch spaces constantly outperforms pure product spaces with same signature. With switch spaces, we are able to achieve the state-of-the-art performances on both tasks, outperforming various recent baselines;  (c) We conduct model analysis to inspect the inner workings of switch spaces and study the effects of certain modeling choices, to provide insights for extension to other applications.


\section{Related Work}

In recent years,  Non-Euclidean geometric representation learning has gained increasing interests. For example,  Hyperbolic representation learning has shown promising performance on a number of tasks such as taxonomic entities modeling~\cite{nickel2017poincare}, network embeddings~\cite{nickel2017poincare}, recommender systems~\cite{vinh2020hyperml}, question answering~\cite{tay2018hyperbolic}, knowledge graphs completion~\cite{balazevic2019multi,chami-etal-2020-low}, graph-related classification~\cite{chami2019hyperbolic,liu2019hyperbolic}, similarity-based hierarchical Clustering~\cite{chami2020trees} and the list goes on. Hyperbolic space is reminiscent of continuous version of trees and excels in modeling hierarchical structures, while spherical space is a more suitable choice for directional similarity modeling and cyclical-structured data. We can find applications of spherical spaces in text embeddings~\cite{meng2019spherical}, texture mapping~\cite{wilson2014spherical}, time-warping functions embedding~\cite{wilson2014spherical}. As indicated, different space has its own specialities  and the choice of spaces varies based on the characteristics of data.

To disburden from space selection, product spaces come into play. A product space is a combination of embedding spaces with heterogeneous curvature (e.g., Euclidean space, hyperbolic space, and spherical space, etc.)~\cite{gu2018learning,skopek2019mixed,bachmann2020constant}. Each component of product space has a constant curvature while the resulting mixed space has non-constant curvature, which makes it possible to capture a wider range of curvatures with lower distortion than single space. Theoretically, product space is well-suited for modeling real-world data with mixture structures. Empirically, product spaces demonstrate its efficacy in graph reconstruction, wording embedding with low dimensions~\cite{gu2018learning}, node classification in graphs~\cite{bachmann2020constant} and image reconstruction~\cite{skopek2019mixed}.

Another line of work that is related this paper is mixture of experts (MOE)~\cite{jacobs1991adaptive,jordan1994hierarchical}. MOE is established based on the divide-and-conquer principle. It divides problem into homogeneous regions and an expert is responsible for each region. The final prediction is arrived based on the cooperation between experts via a gating network. As a controller, the gating networks control the contribution of each experts via a probabilistic gating function. MOE has been extensively studied in a wide spectrum of applications such as language modeling and machine translation~\cite{shazeer2017outrageously,fedus2021switch}, multi-task learning~\cite{ma2018modeling}, etc. Various MOE architectures have also been proposed in the past decades including hierarchical structure~\cite{yao2009hierarchical}, sequential experts~\cite{aljundi2017expert}, deep MOE~\cite{eigen2013learning}, sparsely-gated MOE~\cite{shazeer2017outrageously}, etc.

Different from previous work on product spaces, our work integrate the gating mechanism of MOE~\cite{shazeer2017outrageously} and propose the more expressive switch space, which is a generalization of product space but offers better specialization, higher degree of freedom, and superior computational efficiency.

\section{The Geometry of Product Spaces}

\paragraph{Stereographic Projection model} In general, there are three types of constant curvature spaces with respect to the sign of the curvature. Common realizations are Euclidean space $\mathbb{E}$ (flat), hypersphere $\mathbb{S}$ (positively curved) and hyperboloid $\mathbb{H}$ (negatively curved). For the latter two, we prefer their stereographic projection model: projected sphere $\mathbb{D}$ and Poincare ball $\mathbb{P}$. These models are easier to optimize, avoiding the problem of non-convergence of norm of points with curvature close to $0$, and the projection is conformal, i.e., does not affect the angles between points~\cite{nickel2017poincare,skopek2019mixed}. 

An alternative to vector space in non-Euclidean geometry is gyrovector space~\cite{ungar1991thomas}, which defines operations such as vector addition and multiplication. For $\mathbb{D}$ and $\mathbb{P}$ (jointly denoted as $\mathcal{M}_c$, where $c$ denotes curvature),  the addition between two points $\mathbf{x}, \mathbf{y} \in \mathcal{M}_c$, also knwon as Möbius addition $\oplus_c$ (for both signs of $c$), is defined as:
\begin{equation}
   \mathbf{x} \oplus_c \mathbf{y} = \frac{(1-2c \langle \mathbf{x}, \mathbf{y}\rangle -c \| \mathbf{y} \|^2_2)\mathbf{x} + (1 + c\| \mathbf{x}\|^2)\mathbf{y} }{1 -2c\langle\mathbf{x},\mathbf{y}\rangle + c^2\|\mathbf{x} \|^2_2\| \mathbf{y}\|^2_2},
\end{equation}
Where $\langle,\rangle$ is Euclidean inner product. The distance between points in the gyrovector space is defined as:
\begin{equation}
     d_{\mathcal{M}_c}( \mathbf{x},  \mathbf{y}) = \frac{2}{\sqrt{|c|}}\tan_c^{-1}(\sqrt{|c|} \| - \mathbf{x} \oplus_c  \mathbf{y} \|_2),
\end{equation}
where $\tan_c$ stands for $\tan$ if $c>0$ and $\tanh$ if $c<0$. In both spaces, we have Euclidean geometry when $c \rightarrow 0$. It is easy to prove that: $d_{\mathcal{M}_c}(\mathbf{x}, \mathbf{y}) \xrightarrow{c \rightarrow 0} 2\| \mathbf{x} - \mathbf{y} \|_2$, which means the gyrospace distance converges to Euclidean distance when limiting $c$ to zero.

Let $\mathcal{T}_{\mathbf{x}} \mathcal{M}_c$ be the tangent space to the point $\mathbf{x} \in \mathcal{M}_c$.  Mapping between (Euclidean) tangent space and hyperbolic/spherical space is performed with exponential map: $\mathcal{T}_{\mathbf{x}} \mathcal{M}_c \rightarrow \mathcal{M}_c$ and logarithmic map: $\mathcal{M}_c \rightarrow \mathcal{T}_{\mathbf{x}} \mathcal{M}_c$, which are defined as:
\begin{equation}
\begin{aligned}
  \log_{\mathbf{x}}^c(\mathbf{y}) &= \frac{2}{\sqrt{|c|\lambda_{\mathbf{x}}^c}} \tan_c^{-1} (\sqrt{|c|} \| - \mathbf{x} \oplus_c \mathbf{y} \|_2) \frac{- \mathbf{x} \oplus_c \mathbf{y} }{\| - \mathbf{x} \oplus_c \mathbf{y} \|_2} \\
\exp_{\mathbf{x}}^c(\mathbf{v}) &= \mathbf{x} \oplus_c (\tan_c (\sqrt{|c|}\frac{\lambda_{\mathbf{x}}^c\| \mathbf{v}\|_2}{2})\frac{\mathbf{v}}{\sqrt{|c|}\| \mathbf{v}\|_2}),
\end{aligned}
\end{equation}
where $\lambda_{\mathbf{x}}^c$ is a conformal factor, defined as $\lambda_{\mathbf{x}}^c = 2/(1+c\|\mathbf{x}\|_2^2)$, used to transform the metric tensors between (Euclidean) tangent space and non-Euclidean space.


\paragraph{Product Spaces}
The product space is defined as the Cartesian product of multiple spaces with varying dimensionality and curvature.  Let $\mathcal{P}$ denote a product space which is composed by $N$ independent component space $\mathcal{M}^{(1)}$, $\mathcal{M}^{(2)}$,...,$\mathcal{M}^{(N)}$. The mixed space $\mathcal{P}$ has the following form:
\begin{equation}
    \mathcal{P} = \bigtimes_{i=1}^N \mathcal{M}^{(i)} = \mathcal{M}^{(1)} \times \mathcal{M}^{(2)} \times ... \times \mathcal{M}^{(N)}.
\end{equation}

The product space $\mathcal{P}$ is also equipped with distance function. The squared distances between points $\mathbf{x}, \mathbf{y} \in \mathcal{P}$ is defined as:
\begin{equation}
    d^2_{\mathcal{P}}(\mathbf{x}, \mathbf{y}) = \sum^N_{i=1} d^2_{\mathcal{M}^{(i)}}(\mathbf{x}_{\mathcal{M}^{(i)}}, \mathbf{y}_{\mathcal{M}^{(i)}}),
\end{equation}
where $\mathbf{x}_{\mathcal{M}^{(i)}}$ and $\mathbf{y}_{\mathcal{M}^{(i)}}$ denote the corresponding vectors on the component space $\mathcal{M}^{(i)}$. 

Other operations such as exponential map and logarithmic map are element-wise, which means that we can decompose the points into component spaces and apply operations on each component space and then compose them back (e.g., concatenation) to the product space.

The signature (i.e., parametrization) of product space consists of the types of space, the dimensionality and curvature of each space.





\section{Switch Spaces}
In this section, we motivate and describe the proposed Switch Spaces: a framework for learning product spaces adaptively based on the incoming data. The guiding design principle for Switch Spaces is to endow the learned product spaces model with specialization in a computationally efficient way.


\paragraph{Problem Setup}
Suppose we have $N$ spaces of different types and $\mathcal{M}^{(i)}_c \in \{\mathbb{E}, \mathbb{D}, \mathbb{P}\}$. The dimensions of the spaces are $b_1,b_2,...,b_N$.  The goal of switch spaces is to select $K ( 1 \leq K < N )$ spaces out from the given $N$ spaces for \textit{each} incoming example and make prediction with the selected $K$ ones, making the signature (i.e., the types of component space) of the learned product space switchable. 

Product space model is designed for data with mixed patterns. Yet, it is unclear how to construct a product space model that makes data with certain patterns handled by its suitable space. To calibrate each component space and enhance the alignment between data and its suitable geometries, switch spaces utilize the concept of sparse gating. That is, given a pool of spaces, a sparse gating mechanism is integrated to activate a few spaces for each incoming example, assigning data to its suitable product spaces without manual inspection of the inner structure of input data.

\begin{figure*}[t]
    \centering
    \includegraphics[width=0.8\linewidth]{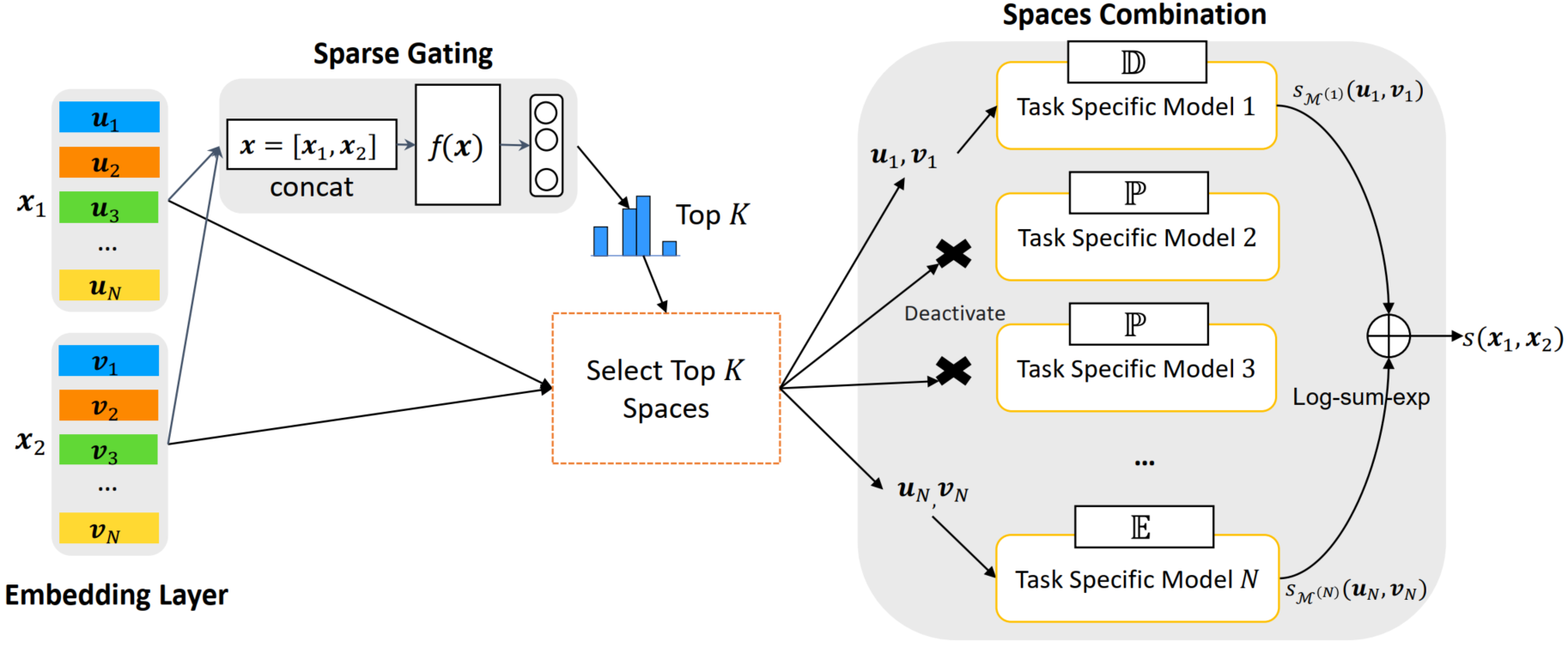}
    \vspace{-0.8em}
    \caption{Architecture of the proposed Switch Spaces. Only $K$ out of $N$ spaces are active for each data point. }
    \label{fig:arch}
\end{figure*}

\paragraph{Embedding Layer}
Let $\mathbf{x}_1=(\mathbf{u}_1,...,\mathbf{u}_N), \mathbf{u}_i \in \mathcal{M}^{(i)}_c$ and $\mathbf{x}_2=(\mathbf{v}_1,...,\mathbf{v}_N), \mathbf{v}_i \in \mathcal{M}^{(i)}_c$  denote two points on the switch space. In real world applications, $\mathbf{x}_1$ and $\mathbf{x}_2$ can represent different meanings (e.g., users/items in recommender systems, entities/relations in knowledge graphs) and sometimes more than two vectors are needed. For simplicity, we do not distinguish them in this section. We initialize all the embedding vectors $\mathbf{x}_1$ and $\mathbf{x}_2$ in tangent space which is beneficial in two-folds: On the one hand, the embedding layer will be further used for the sparse gating network in Euclidean space. On the other, standard Euclidean optimization techniques can be easily applied thanks to the convenient exponential and logarithmic maps. As such, the overall framework can be optimised in a uniform fashion, avoiding the cumbersome Riemannian optimization~\cite{bonnabel2013stochastic}. Exponential map will be applied to recover the hyperbolic/spherical parameters when necessary.

\paragraph{Selecting Top-K Manifolds}
The sparse gating network is a differentiable data-driven tool to select the most suitable $K$ spaces for each sample.


We provide two options to construct the gating network which mainly differ in the input and hidden layers. The first option is to construct the input layer by concatenating all the embedding vectors to obtain a one-dimensional vector with length $2\sum_{i}^N b_i$:
\begin{equation}
    \mathbf{x} = [\mathbf{x}_1, \mathbf{x}_2] = [\mathbf{u}_1,...,\mathbf{u}_N,\mathbf{v}_1,...,\mathbf{v}_N].
\end{equation}
Then linear transformations or 1-D convolutions with linear transformations can be applied over $\mathbf{x}$ to transform it into an $N$ dimensional vector.

Another option is only viable when all embedding vectors have the same dimensionality ($b_1=...b_i=...b_N$). We concatenate them into a two-dimensional matrix with shape $2N \times b_i $. Then 2-D convolutions with linear transformations can be applied. Using convolutional operations can in general reduce the number of parameters when $b_i$ is large.

In formal, the last hidden layer of the gating network has the following form:
\begin{equation}
     f(\mathbf{x}) = f_1(\mathbf{x}) + \text{randn}() \cdot \ln(1 + \exp(f_2(\mathbf{x}) )),
\end{equation}
where $f_1$ and $f_2$ represent linear layers or CNN layers and $f_*(\mathbf{x}) \in \mathrm{R}^N$; function ``$\text{randn}()$" creates samples from the standard normal distribution; the second term is a tunable Gaussian noise (ony for training) that is added to improve the load balancing, i.e., balance the number of samples accepted by each space. 





To impose sparsity, we employ a $\text{TopK}$ function to obtain the largest $K$ elements from $f(\textbf{x})$:
\begin{equation}
    f(\mathbf{x}) \leftarrow \text{TopK}(f(\mathbf{x})),
\end{equation}
where the $\text{TopK}$ function returns the original value if the element is in the top $K$ list, otherwise it returns $-\infty$.

We need $N$ gates $g_1, g_2,..,g_N$ to generate probabilities to control the activeness of each space. 
\begin{equation}
    0 \leq g_1(\mathbf{x}), g_2(\mathbf{x}),...g_N(\mathbf{x}) \leq 1, \sum_{i=1}^N g_i(\mathbf{x}) = 1.
\end{equation}
Simply, a softmax gating network is employed. For $-\infty$, it will output a zero gate value, which introduces sparsity to the gates.
\begin{equation}
    g_i(\mathbf{x}) = \frac{\exp(f(\mathbf{x})_i)}{\sum_{j=1}^N\exp(f(\mathbf{x})_j)}.
\end{equation}

Only spaces with nonzero gate values will be active while other spaces are idle and not computed. In doing so, it provides the possibility to enlarge the model size without incurring much computational cost as long as the number of active spaces $K$ is fixed.

\paragraph{Task Specific Models}
The model structures are associated with the type of tasks and spaces. In each component space,  the input vectors $\mathbf{u}_i$ and $\mathbf{v}_i$ are processed with corresponding exponential/logarithm map for further operations like Mobius addition and distancing function. We elaborate the task specific model in the following section. Let $s_{\mathcal{M}^{(i)}}(\mathbf{u}_i, \mathbf{v}_i)$ (e.g., squared distance, etc.) denote the output of the $i^{\text{th}}$ model. We set $s_{\mathcal{M}^{(i)}}$  to $0$ if the corresponding space is inactive.

\paragraph{Scoring Function of Switch Space}
Similar to the definition of squared distance on product space, the scoring function of switch space also decomposes and we defined it as:
\begin{equation}
\begin{aligned}
s(\mathbf{x}_1, \mathbf{x}_2) &=\log( \sum_{i=1}^N  g_i(\mathbf{x}) \exp(s_{\mathcal{M}^{(i)}}(\mathbf{u}_i, \mathbf{v}_i)) ).
\end{aligned}
\end{equation}
A log-sum-exp technique is adopted to improve the numerical stability, avoiding problems such as underflow and overflow. Note that multiplying by the gating probability is optional in our framework. If $s_{\mathcal{M}^{(i)}}$ is calculated with squared distance and $K=N$, we can recover the squared distance function defined for product space by removing $g_i(\mathbf{x})$ and the log-sum-exp technique. Our scoring function is not a standard distance metric as it does not satisfy the triangle inequality since we cannot guarantee that two pairs of points have the same active component spaces. In addition, scoring functions other than distance can also be used to allow more complex task-specific models. Relaxing these constraints gives the framework more flexibility.



The architecture of switch space is shown in Figure \ref{fig:arch}. The whole framework can be optimized end-to-end with model specific loss functions. Also, we add an additional loss function as that discussed in~\cite{shazeer2017outrageously} for load balancing.

\paragraph{Discussion}

Unlike usual product spaces where the signature is fixed, the signature of the resulting product space in switch spaces may varying according to the gating values. Theoretically, there will be $C_N^K = \frac{N!}{(N -K)! K!}$ possible combinations of product spaces in the switch spaces, allowing greater degree of freedom.

In term of computational complexity, our model has a constant computational cost pertaining to $K$, regardless of the model sizes (number of model parameters).

In product space, product of $\mathbb{E}^{b_1},...,\mathbb{E}^{b_N}$ is identical to the single space $\mathbb{E}^{b_1+...+b_N}$. However, this is not the case in switch space when $K<N$ because the active Euclidean spaces vary.

\section{Experiments}
In this section, we build models for knowledge graph completions and item recommendations utilizing the proposed switch spaces.  Both tasks heavily relies on representation learning. we firstly generalize models into hyperbolic and spherical spaces (if not existed) and then construct the product space and switch space counterparts. We conduct experiments on four real-world datasets and compare with recent proposed baselines.  All our experiments are run on NVIDIA TITAN Xp GPUs.

\subsection{Knowledge Graph Completion}
Knowledge graphs (KG) are fundamental tools for representing information and have applications in wide range of tasks such as question answering and web search. Knowledge graphs are highly incomplete, making it challenging for downstream use. Here, the goal is to infer the missing facts in knowledge graph by embedding the entities and relations into appropriate spaces.

\subsubsection{Model Details} Given a knowledge graph $\mathcal{G}$ with a set of entities $\mathcal{E}$ and a set of relations $\mathcal{R}$. Each triplet, abbreviated as $(h, r, t)$, in $\mathcal{G}$ is composed by two entities (i.e the head entity $h \in \mathcal{E}$ and the tail entity $t \in \mathcal{E}$), and the relationship $r \in \mathcal{R}$ between them. 

We follow the base model RotE (Euclidean) and RotH (Poincare)~\cite{chami-etal-2020-low} and propose a spherical model. In formal, entities $h, t$ is represented by vectors $\mathbf{e}_h, \mathbf{e}_t \in \mathbb{R}^b$ and relation $r$ is represented by two translation vectors $\mathbf{\alpha}_r, \mathbf{\beta}_r \in \mathbb{R}^b$ and a rotation vector $\mathbf{\gamma}_r \in \mathbb{R}^b$. The head entity is translated twice via Mobius addition in the spherical spaces and rotated once:
\begin{equation}
    Q(h, r)  = \text{Rot}(\exp_{\mathbf{0}}^c(\mathbf{e}_h) \oplus_c \exp_{\mathbf{0}}^c(\mathbf{\alpha}_r), \mathbf{\gamma}_r) \oplus_c \exp_{\mathbf{0}}^c(\mathbf{\beta}_r),
\end{equation}
where $c>0$ and $\exp_{\mathbf{0}}^c$ is the exponential map over origin. $\text{Rot}$ is a rotation function and $ \mathbf{\gamma}_r$ is the rotation matrix. 

It is then compared with the tail entities using squared spherical distance. The final scoring function is as follows:
\begin{equation}
     s(h, r, t)  = -d_{\mathcal{M}_c}^2(Q(h, r), \exp_{\mathbf{0}}^c(\mathbf{e}_t)) + b_h + b_t,
\end{equation}
where $b_h, b_t \in \mathbb{R}$ are head and tail specific biases and are trainable parameters. Same as RotH, we make the curvature relation specific and trainable, i.e, each relation has a curvature parameter and are updated simultaneously with other parameter. Interested readers are referred to ~\cite{chami-etal-2020-low} for more details.





We use the above model together with RotE, RotH as the task-specific models for switch space and train the whole framework by minimizing the following cross-entropy loss.
\begin{equation}
    \mathcal{L} = \sum_{(h,r,t) \in \Omega} \log(1+ \exp(-Y_{(h,r,t)} s(h,r,t))) ,
\end{equation}
where $Y_{(h,r,t)} \in \{1, -1\}$ is a binary label indicating whether a triplet is factual (1) or not (-1). $\Omega$ represents the training collection which includes both positive and negative triplets.

For simplicity, we assume that the value  of $b$ is identical for all spaces. We concatenate $\mathbf{e}_h$ with $\alpha_r$ and $\beta_r$ of all spaces and obtain a two dimensional matrix of size $3N \times b$, which is input to the gating network (using 2-D CNN) of switch space. We denote our approach as \textbf{SwisE} ( short for \textbf{Swi}tch \textbf{S}pace based KG \textbf{E}mbeddings).

\begin{table}[t]
\centering
\small
\begin{tabular}{cccccc}
\toprule
Dataset   & $|\mathcal{E}|$ & $|\mathcal{R}|$ & \#training & \#val & \#test  \\
\midrule
WN18RR    &   40943 & 11  &       86835     &     3034         &    3134  \\
FB15K-237 &  14541 & 237  &    272115        &   17535           &      20466 \\ \bottomrule
\end{tabular}
\caption{Statistics of the WN18RR and FB15K-237 datasets.}
\label{kgdata}
\end{table}

\begin{table*}[t]
\centering
\begin{small}
\begin{tabular}{
M{5.8em} L{5.8em}
M{2.8em} M{2.8em} M{2.8em} M{2.8em}
M{0em}
M{2.8em} M{2.8em} M{2.8em} M{2.8em}
}
\toprule
\multirow{2}{*}{Space} & \multirow{2}{*}{Model} & 
\multicolumn{4}{c}{WN18RR} & \phantom{}  & \multicolumn{4}{c}{FB15K-237} \\ 
\cmidrule{3-6} \cmidrule{8-11} 
 && \multicolumn{1}{c}{MRR} &\multicolumn{1}{c}{HR@1} & \multicolumn{1}{c}{HR@3} & \multicolumn{1}{c}{HR@10} && \multicolumn{1}{c}{MRR} & \multicolumn{1}{c}{HR@1} & \multicolumn{1}{c}{HR@3} & \multicolumn{1}{c}{HR@10} \\ \midrule

 \multirow{5}{*}{$\mathbb{R}$} & TransE  &  0.226 & -& - & 0.501 && 0.294& -& -& 0.465\\  
&  BoxE  &  0.451 & -& - & 0.541 && 0.337& -& -& 0.538\\ 
& DistMult  &  0.430 & 0.390& 0.440 & 0.490 && 0.241& 0.155& 0.263& 0.419 \\ 
& ConvE  &  0.430 & 0.400& 0.440 & 0.520 && 0.325& 0.237& 0.356& 0.501 \\
& TuckER  &  0.470 & 0.443& 0.482 & 0.526 && 0.358& 0.266& 0.394& 0.544 \\ 
& RotE&  0.494 & 0.446& 0.512  & 0.585 && 0.346   &   0.251 & 0.381   & 0.538\\ \midrule

\multirow{2}{*}{$\mathbb{C}$} & ComplEx-N3   & 0.480& 0.435& 0.495& 0.572 && 0.357& 0.264& 0.392& 0.547 \\ 
 & RotatE  & 0.476& 0.428& 0.492& 0.571 && 0.338& 0.241& 0.375& 0.533 \\ \midrule

$\mathbb{Q}$ & QuatE & 0.488& 0.438& 0.508& 0.582 &&\underline{ 0.366  } &\underline{ 0.271 }  & \underline{0.401  } & \underline{0.556} \\ \midrule

\multirow{2}{*}{$\mathbb{P}$} & MurP  &  0.481 & 0.440& 0.495& 0.566 && 0.335& 0.243& 0.367& 0.518 \\ 
& RotH & \underline{0.496 } & \underline{0.449}& \underline{0.514  } & \underline{0.586} &&0.344 & 0.246& 0.380& 0.535 \\ \midrule

Switch Spaces                 &      \textbf{SwisE}          & \textbf{0.526} & \textbf{ 0.480} & \textbf{0.549} &\textbf{  0.610} & & \textbf{0.521} &\textbf{0.486} & \textbf{0.530} &\textbf{ 0.585} \\ \bottomrule

\end{tabular}
\caption{Comparison with state-of-the-arts on datasets WN18RR and FB15K-237. The best performances are in boldface and the second best are underlined. The performance of switch space on WN18RR is achieved with model $(\mathbb{D}^{100})^4\mathbb{E}^{100}$(K=2), and that on FB15K-237 is with $(\mathbb{P}^{100})^3(\mathbb{D}^{100})^2(K=4)$. $\mathbb{R}$:real coordinate space; $\mathbb{C}$:complex number space; $\mathbb{Q}$:quaternion space.}
\label{tab:kbc}
\end{small}
\end{table*}

\subsubsection{Experimental Setup}

We use two standard datasets including WN18RR~\cite{bordes2013translating,dettmers2018convolutional} and FB15K-237~\cite{bordes2013translating,dettmers2018convolutional}. WN18RR is taken from  WordNet, a lexical database of semantic relations between words. FB15K-237 is a subset of the Freebase knowledge graph which is a global resource consisting of common and general information. Statistics of the two datasets is shown in Table \ref{kgdata}.


We compare our model with Euclidean methods TransE~\cite{bordes2013translating}, DistMult~\cite{yang2014embedding}, ConvE~\cite{dettmers2018convolutional}, TuckER~\cite{balazevic2019tucker}, RotE~\cite{chami-etal-2020-low}, and BoxE~\cite{abboud2020boxe}; complex number based methods ComplEx-N3,~\cite{lacroix2018canonical} and RotatE~\cite{sun2018rotate}; quaternion model QuatE~\cite{zhang2019quaternion}; and hyperbolic methods MurP~\cite{balazevic2019multi} and RotH~\cite{chami-etal-2020-low}.

The performance of different models are evaluated using two standard metrics including mean reciprocal rank (MRR) and hit rate (HR) with given cut-off value $\{1, 3, 10\}$. 


The total dimension is fixed to $500$ for fair comparison. Learning rate is tuned among $\{0.01, 0.005, 0.001\}$. For all experiments, we reports the average over $5$ runs. We set the kernel size to $5$ and stride to $3$ for convolution operation in the gating network. $N$ is set to $5$ and $K$ is tuned among  $\{1, 2, 3, 4\}$. The number of negative samples size per factual triplet is set to $50$. Optimizer Adam is used for model learning.

\begin{table}[t]
\small
\centering
\begin{tabular}{l l c c}
\toprule
Space                         & Model             & MRR               & HR@3            \\ 
\midrule
Single &  $\mathbb{D}^{500}$       &0.492 & 0.514\\ 
\midrule
\multirow{6}{*}{Product}                  &          $(\mathbb{P}^{100})^5$          &      0.483        &               0.498 \\ 
 &          $(\mathbb{D}^{100})^5$          &         0.480       &          0.506        \\ 
 &        $(\mathbb{P}^{100})^3(\mathbb{D}^{100})^2$            &    0.484             &  0.498                \\ 
 &        $(\mathbb{D}^{100})^4\mathbb{E}^{100}$            &      0.479           & 0.497                 \\ 
&        $(\mathbb{P}^{100})^4\mathbb{E}^{100}$            &       0.477       &        0.498         \\

&        $(\mathbb{P}^{100})^2(\mathbb{D}^{100})^2\mathbb{E}^{100}$            &        0.479        &       0.496           \\
\midrule
\multirow{6}{*}{\begin{tabular}[c]{@{}c@{}}Switch\\  (K=2)\end{tabular}}       
 &        $(\mathbb{P}^{100})^5$            &     0.516            &          0.539      \\ 
 &        $(\mathbb{D}^{100})^5$            &     0.509             &          0.531      \\ 
 &        $(\mathbb{P}^{100})^3(\mathbb{D}^{100})^2$            &        0.500            &     0.521        \\ 
 &        $(\mathbb{D}^{100})^4\mathbb{E}^{100}$            &       0.526          &    0.549                   \\ 
&        $(\mathbb{P}^{100})^4\mathbb{E}^{100}$            &        0.517     &       0.542        \\
&        $(\mathbb{P}^{100})^2(\mathbb{D}^{100})^2\mathbb{E}^{100}$            &    0.522           &    0.546             \\
\bottomrule
\end{tabular}
\caption{Comparison with Product Spaces Models (Only representative signatures are reported due to length constraints).}
\label{tab:product}
\vspace{-1em}
\end{table}

\begin{table*}[t]
\begin{small}
\small
\centering
\begin{tabular}{
M{4.8em} L{6.2em}
M{5.em} M{5.0em} M{5em} 
M{0em}
M{5em} M{5em} M{5em} 
}
\toprule
\multirow{2}{*}{Space} & \multirow{2}{*}{Model} & \multicolumn{3}{c}{MovieLens 100K}          & \phantom{} &  \multicolumn{3}{c}{MovieLens 1M}               \\ \cmidrule{3-5} \cmidrule{7-9} 
&  & MAP       & P@10 & R@10 & & Map       & P@10 & R@10  \\ \midrule
   \multirow{4}{*}{\begin{tabular}[c]{@{}c@{}}Single\end{tabular}}  &     BPRMF                   &       .211$\pm$.003  &     .254$\pm$.001  &      .176$\pm$.004  &  &    .120$\pm$.001      &     .185$\pm$.001       &      .078$\pm$.001      \\ 
 &     $\mathbb{E}^{100}$(CML)    &      .195$\pm$.001 &       .233$\pm$.003    &       .165$\pm$.003  &   &    .138$\pm$.002    &   .208$\pm$.002                   &  .092$\pm$.003     \\ 
 & $\mathbb{P}^{100}$(HyperML)   &   .157$\pm$.001    &    .194$\pm$.003     &    .136$\pm$.004  &  &      .113$\pm$.001 &       .167$\pm$.002    &     .075$\pm$.002   \\
& $\mathbb{D}^{100}$     &     .161$\pm$.001  &   .193$\pm$.002      &     .137$\pm$.001  &  &   .116$\pm$.001  &   .170$\pm$.004     &    .077$\pm$.001                \\ \midrule
\multirow{6}{*}{\begin{tabular}[c]{@{}c@{}}Product\end{tabular}}          
&   $(\mathbb{E}^{20})^5$              &    .199$\pm$.002  &  .240$\pm$.001     &     .168$\pm$.003 &  &     .150$\pm$.001    &     .221$\pm$.002       &   .100$\pm$.001    \\        
&   $(\mathbb{P}^{20})^5$              &    .169$\pm$.003  &  .210$\pm$.004    &  .146$\pm$.003 &  &   .128$\pm$.003  & .189$\pm$.002&    .083$\pm$.001\\     

& $(\mathbb{D}^{20})^5$                    &      	.177$\pm$.003   &  	.212$\pm$.002      & 	.149$\pm$.005      &   &  .131$\pm$.001 &    .188$\pm$.001   &  .085$\pm$.004    \\ 

& $(\mathbb{D}^{20})^3(\mathbb{P}^{20})^2$                    &     .173$\pm$.001  	  &  .209$\pm$.001	      & .151$\pm$.001	 &  &  .133$\pm$.001 & .193$\pm$.001   &    .085$\pm$.001      \\ 
                      
&     $(\mathbb{P}^{20})^2(\mathbb{D}^{20})^2\mathbb{E}^{20} $                 & .176$\pm$.001 &  .208$\pm$.002 & .149$\pm$.003  &    & 	.134$\pm$.002 & .192$\pm$.003 & .088$\pm$.002\\
\midrule
\multirow{6}{*}{\begin{tabular}[c]{@{}c@{}}Switch\\ (K=4)\end{tabular}} 
&       $(\mathbb{E}^{20})^5 $        &  \textbf{.218}$\pm$.003  &   \textbf{.267}$\pm$.003      &   \textbf{.188}$\pm$.002 &    & .158$\pm$.001 & .236$\pm$.001  &  .105$\pm$.002 \\ 
&     $(\mathbb{P}^{20})^5$                 & .191$\pm$.002 &  .233$\pm$.005  &  .162$\pm$.004     &  & .142$\pm$.003 & .213$\pm$.005 &.094$\pm$.005    \\ 

&     $(\mathbb{D}^{20})^5$                 & .194$\pm$.002 & .233$\pm$.005   &      .162$\pm$.002 &  & .172$\pm$.003& \textbf{.253}$\pm$.003    & \textbf{.116}$\pm$.002  \\

& $(\mathbb{D}^{20})^3(\mathbb{P}^{20})^2$                    &     .203$\pm$.003	  &  .246$\pm$.002	  & .174$\pm$.003    & 	  & \textbf{.174}$\pm$.001 & .252$\pm$.002 & \textbf{.116}$\pm$.001\\

&     $(\mathbb{P}^{20})^2(\mathbb{D}^{20})^2\mathbb{E}^{20} $                 & .201$\pm$.002&   .245$\pm$.004 &.171$\pm$.001 && .172$\pm$.001 & .249$\pm$.002 & .114$\pm$.002\\
\bottomrule
\end{tabular}
\caption{ Recommendation performances on two benchmark datasets.  Only representative signatures are reported due to length constraints.}
\label{tab:recsys}
\end{small}
\end{table*}

\subsubsection{Results}

\paragraph{Comparison with the State-of-the-art} Table \ref{tab:kbc} compares our best scores with baselines. Notably, SwisE achieves the new state-of-the-art results across all metrics on both datasets. SwisE outperforms Euclidean, hyperbolic, complex-valued approaches, and obtains a clear performance gain over the second best baseline. Specifically,  the improvement on FB15K-237 is quite impressive, this is because that FB15K-237 has a more diverse set of relations than WN18RR~\cite{balazevic2019multi}, which is also the reason why RotH is surpassed by several Euclidean/complex-valued baselines. Additionally, we observe that RotH and QuatE achieve the second best on WN18RR and FB15K-237, respectively.

\paragraph{Comparison with Product Space}
We compare the switch spaces with product spaces with varying signatures in Table \ref{tab:product} on WN18RR.  We observe that switch spaces constantly outperforms pure product spaces with the same prior signatures, which ascertains the effectiveness of switch spaces. It is worth noting that SwisE requires merely two active spaces for each instance to outperforms product spaces with five spaces. We find that SwisE with the product of spherical and Euclidean spaces archives better performances than other configurations.

\begin{table}[t]
\small
\centering
\begin{tabular}{lcccc}
\toprule
Dataset   & $C_1$ & $C_2$ & \#Interactions & density \\
\midrule
MovieLens 100K    &   943 & 1682  &       100,000     &     0.063        \\
MovieLens 1M &  6040 & 3706  &    1,000,209        &   0.045            \\ \bottomrule
\end{tabular}
\caption{Statistics of MovieLens 100K and MovieLens 1M.}
\label{data:recsys}
\end{table}

\subsection{Item Recommendation} 

 Recommender system is an important part in modern e-commerce platforms. It can improve customers experience via reducing information overload, and increase revenue for companies. Learning representations that reflect users' preference and items characteristics has been a central theme in item recommendations research~\cite{ricci2011introduction}. 

\subsubsection{Model Details} 
Given a collection of $C_1$ users and $C_2$ items and interactions observed between them, the goal is to produce personalized items recommendation for each user.  We are interested in metric learning based collaborative filtering~\cite{hsieh2017collaborative,vinh2020hyperml}. Taken a user vector $\mathbf{u}_u, \mathbf{u} \in \mathcal{M}_c^{C_1 \times b}$ and an item vector $\mathbf{v}_i, \mathbf{v} \in \mathcal{M}_c^{C_2 \times b}$ as input. The the preference of user $u$ towards item $i$ is measured by the squared distance between them:
\begin{equation}
    s(u, i) = -d^2_{\mathcal{M}_c}(\mathbf{u}_u, \mathbf{v}_i).
\end{equation}
We optimize the model with a max-margin hinge loss. 
\begin{equation}
    \mathcal{L} = \sum_{(u,i, j) \in \Omega} \max (0, s(u,i) + m - s(u,j) ),
\end{equation}
where $\Omega$ is the collection of training samples; $i$ is the item that $u$ liked and $j$ represents a negative (unobserved) item for the user. $\max(0,x)$ is also known as the \textit{ReLU} function. $m$ is a margin value.

We also use the same $b$ for all component spaces for brevity. The input of the gating network is the concatenation of the user and item embeddings. Specifically, a one dimensional vector of size $\mathrm{R}^{2Nb} $ is used as the input and a linear layer is applied in the gating network.

\subsubsection{Experimental Setup}
We conduct our experiments on two datasets: MovieLens 100K and MovieLens 1M~\cite{harper2015movielens}. We hold  70\% actions in each user's interactions as the training set, 10\% actions as the validation set for model tuning and the remaining 20\% actions as the test set.  All interactions (e.g., ratings) are binarized following the implicit feedback setting~\cite{rendle2009bpr}. We estimate the global average curvature with the algorithm described in \cite{gu2018learning} for the two datasets and obtain $0.190$ for MovieLens 100K and $0.695$ for MovieLens 1M, which suggests that they lean towards Euclidean/cyclical structures. Statistics of them are in Table \ref{data:recsys}.

We compare our method with a number of baselines including Bayesian personalized ranking based matrix factorization (BPRMF)~\cite{rendle2009bpr}, Euclidean, hyperbolic, spherical, and product space models. $\mathbb{E}^n$ corresponds to CML~\cite{hsieh2017collaborative}, $\mathbb{P}^n$ is equivalent to HyperML~\cite{vinh2020hyperml}.

We measure the performance based on the widely adopted metrics in recommender systems: mean average precision (MAP), precision ($P@5$ and $P@10$), and recall ($R@5$ and $R@10$). 


For all models, the total dimension is fixed to $100$ for fair comparison. The curvatures for spherical and hyperbolic models are set to $1$ and $-1$, respectively. $N$ is set to $5$ and $K$ is tuned among  $\{1, 2, 3, 4\}$. Regularization rate is chosen from $\{0.1, 0.01, 0.001\}$. $m$ is fixed to $0.5$. Adam is also adopted as the optimizer.

\begin{figure*}[t]
\begin{center}
\begin{minipage}[r]{6.5cm}
\includegraphics[width=1\linewidth]{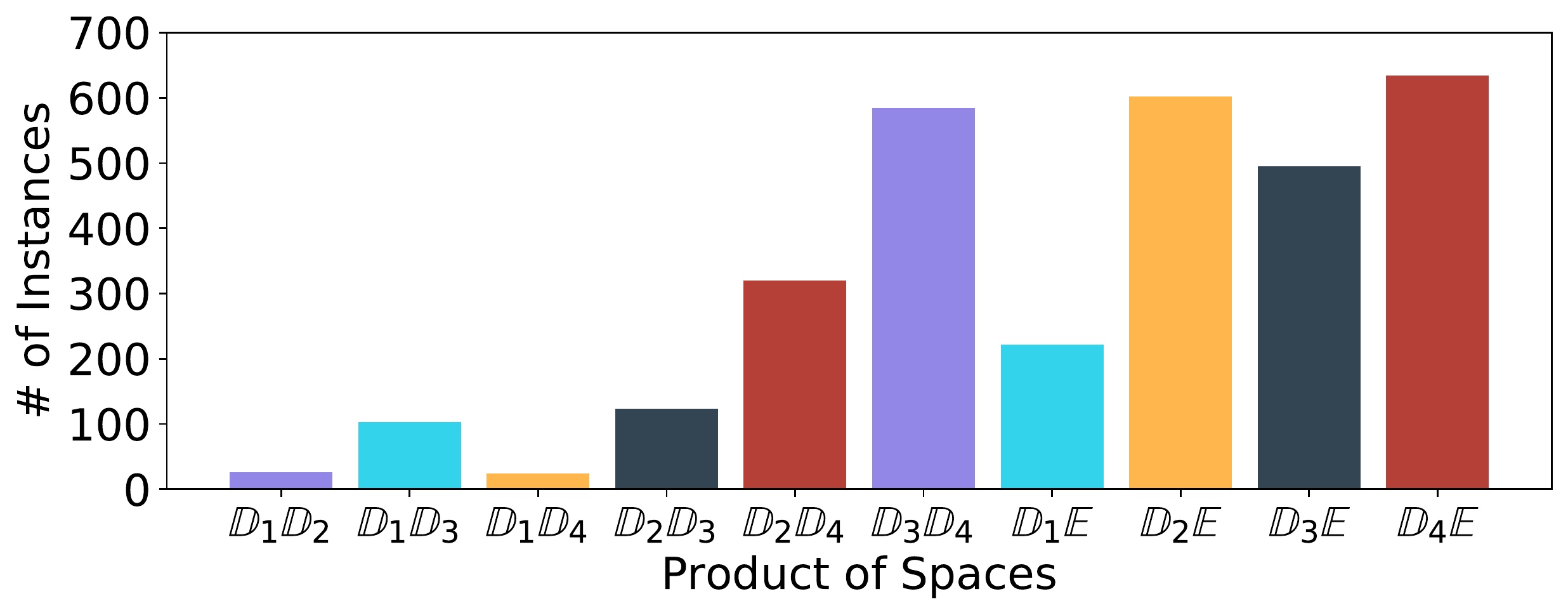}
\end{minipage}
\begin{minipage}[r]{6.5cm}
\includegraphics[width=1\linewidth]{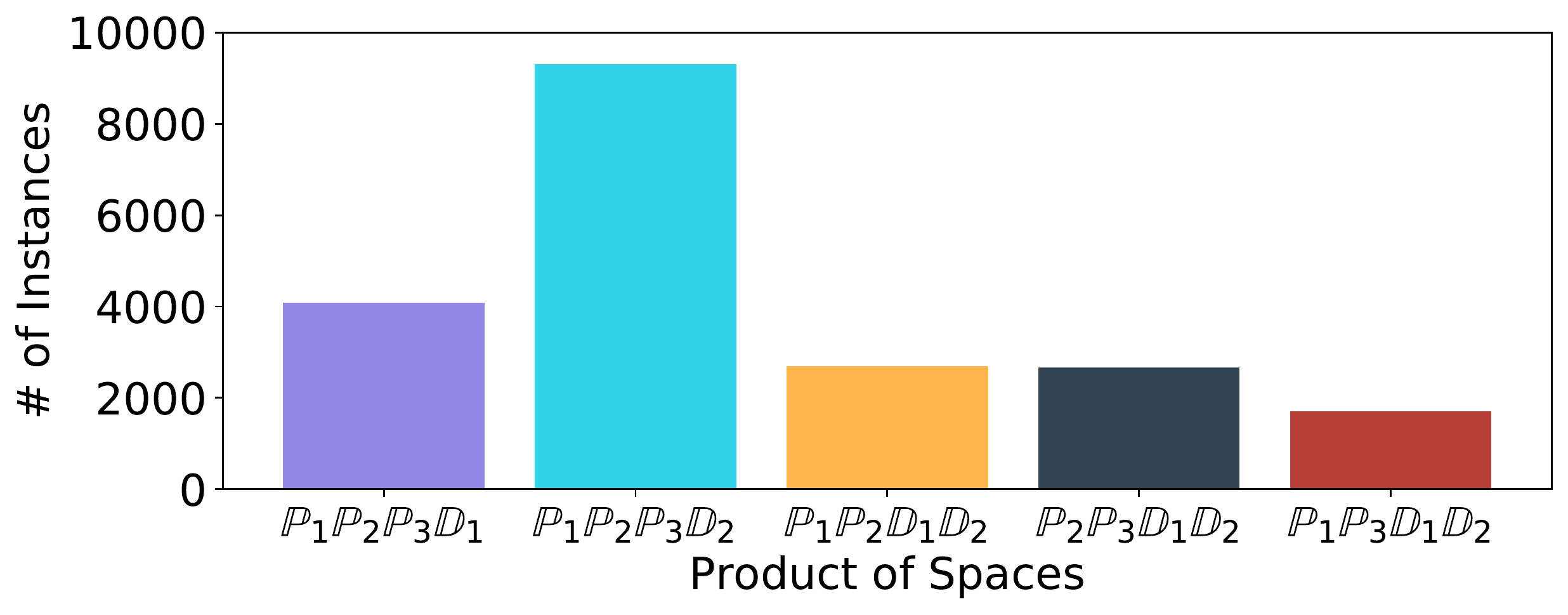}
\end{minipage}
\vspace{-0.8em}
\caption{Distribution of the learned spaces on test data of WN118RR (left figure, $(\mathbb{D}^{100})^4\mathbb{E}^{100}(K=2)$) and FB15K-237 (right figure, $(\mathbb{P}^{100})^3(\mathbb{D}^{100})^2(K=4)$) using switch space.}
\vspace{-0.8em}
\label{fig:distrib}
\end{center}
\end{figure*}

\subsubsection{Results}

Table \ref{tab:recsys} summarizes the performances of our model. The merits of switch spaces can be further observed from the comparison. On both datasets, switch spaces based models achieve the best performances, outperforming their pure product space counterparts with the same signature. Another clear pattern is that product spaces usually performs better than single spaces. Specifically, on MovieLens 100K, Euclidean space is the most effective. While, on MovieLens 1M, spherical space dominated models outperform other space related models. This observation is consistent with the global average curvature we estimated. Interestingly, we find that the matrix factorization approach (BPRMF) remains competitive on the two datasets and usually achieves better results than other single space models.

\section{Model Analysis}

\subsection{Distribution of the Learned Spaces}

\begin{figure}[h]
    \centering
    \includegraphics[width=0.75\linewidth]{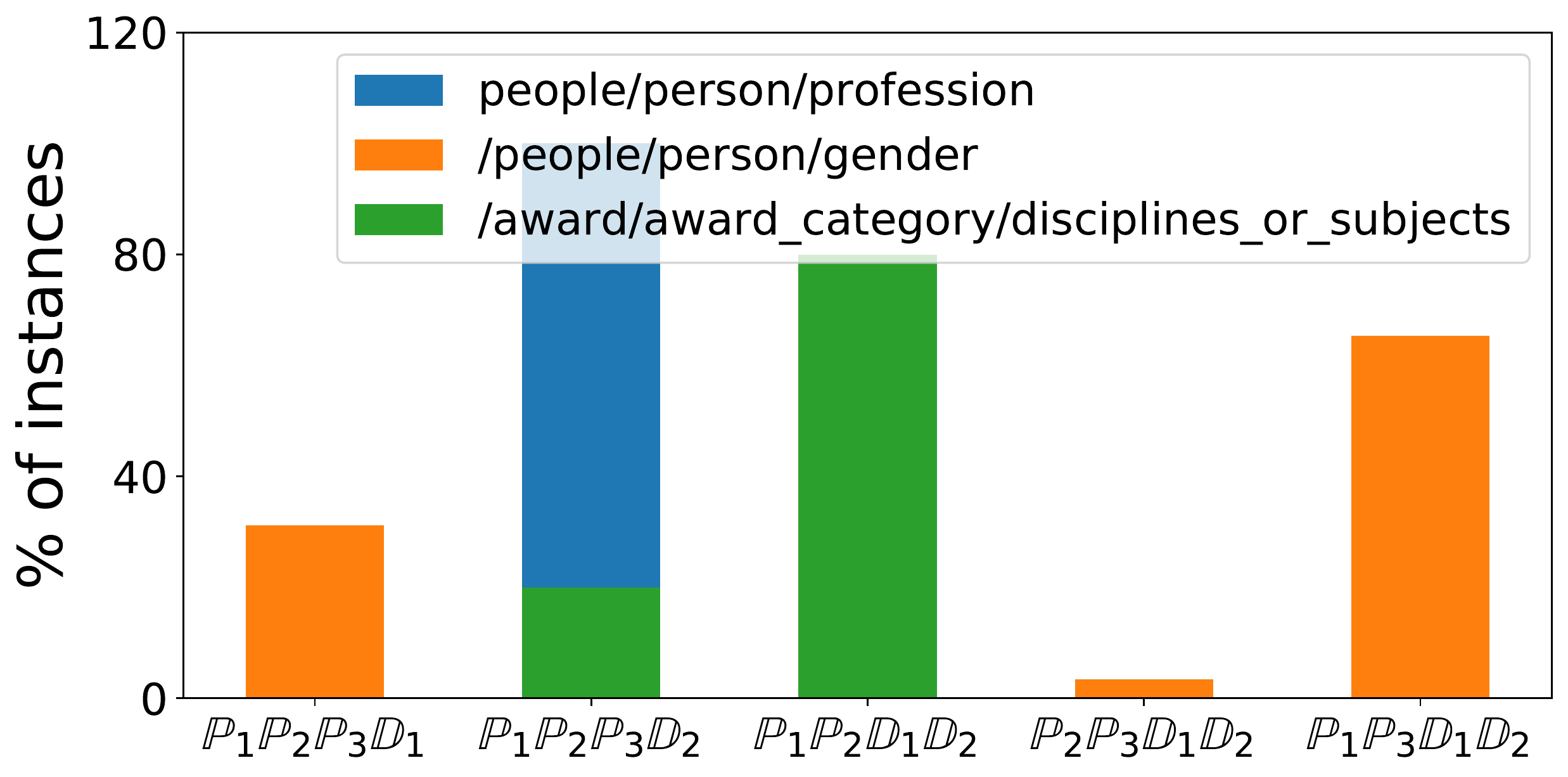}
    \vspace{-0.8em}
    \caption{Distribution of the learned spaces of three examples from FB15K-237. Counts of instances are normalized.}
    \label{fig:samplecase}
\vspace{-1em}
\end{figure}

Figure \ref{fig:distrib} presents the distribution of the learned spaces combinations. It is obvious that samples are not randomly distributed. Instead, certain products are preferred in each case. For example, the product between spherical and Euclidean spaces are more popular on WN18RR. While on FB15K-237, hyperbolic space dominated spaces are preferable. To show more fine-grained examples,  we randomly select three relations from FB15K-237. The distribution of samples that contain these relations are shown in Figure \ref{fig:samplecase}. We find that relations have their own desirable product spaces as well. This reconfirms the specification capability of the proposed switch spaces.

\subsection{Effects of Total Spaces Number}

\begin{figure}[H]
    \centering
    \includegraphics[width=0.65\linewidth]{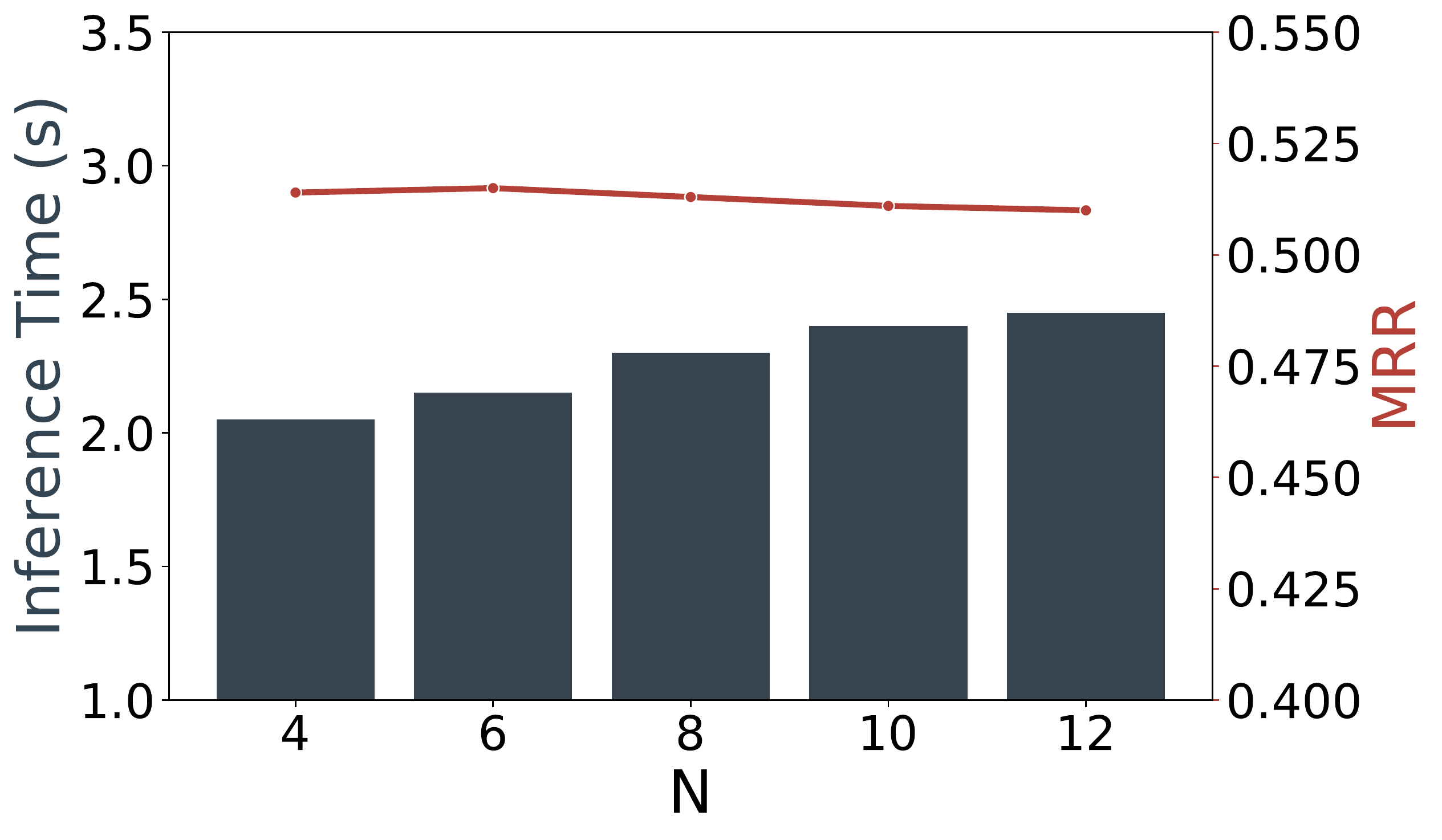}
    \vspace{-0.8em}
    \caption{Effect of $N$ with model $(\mathbb{P}^{100})^N$ (K=2) on MRR and inference time, on dataset WN18RR.}
    \label{fig:n}
\vspace{-1em}
\end{figure}

Figure \ref{fig:n} reports the test MRR and inference time varied across the different $N$ values with $K=2$ on dataset WN18RR using switch space model $(\mathbb{P}^{100})^N$. We observe that, in this task, the model performance does not benefit from the increasing model size. Also, the inference time remains nearly constant when we increase $N$, confirming the efficiency of switch spaces. This property is important for tasks (e.g., language modeling) that prefer larger model size.

\subsection{Effects of Active Spaces Number }

\begin{figure}[H]
    \centering
    \includegraphics[width=0.65\linewidth]{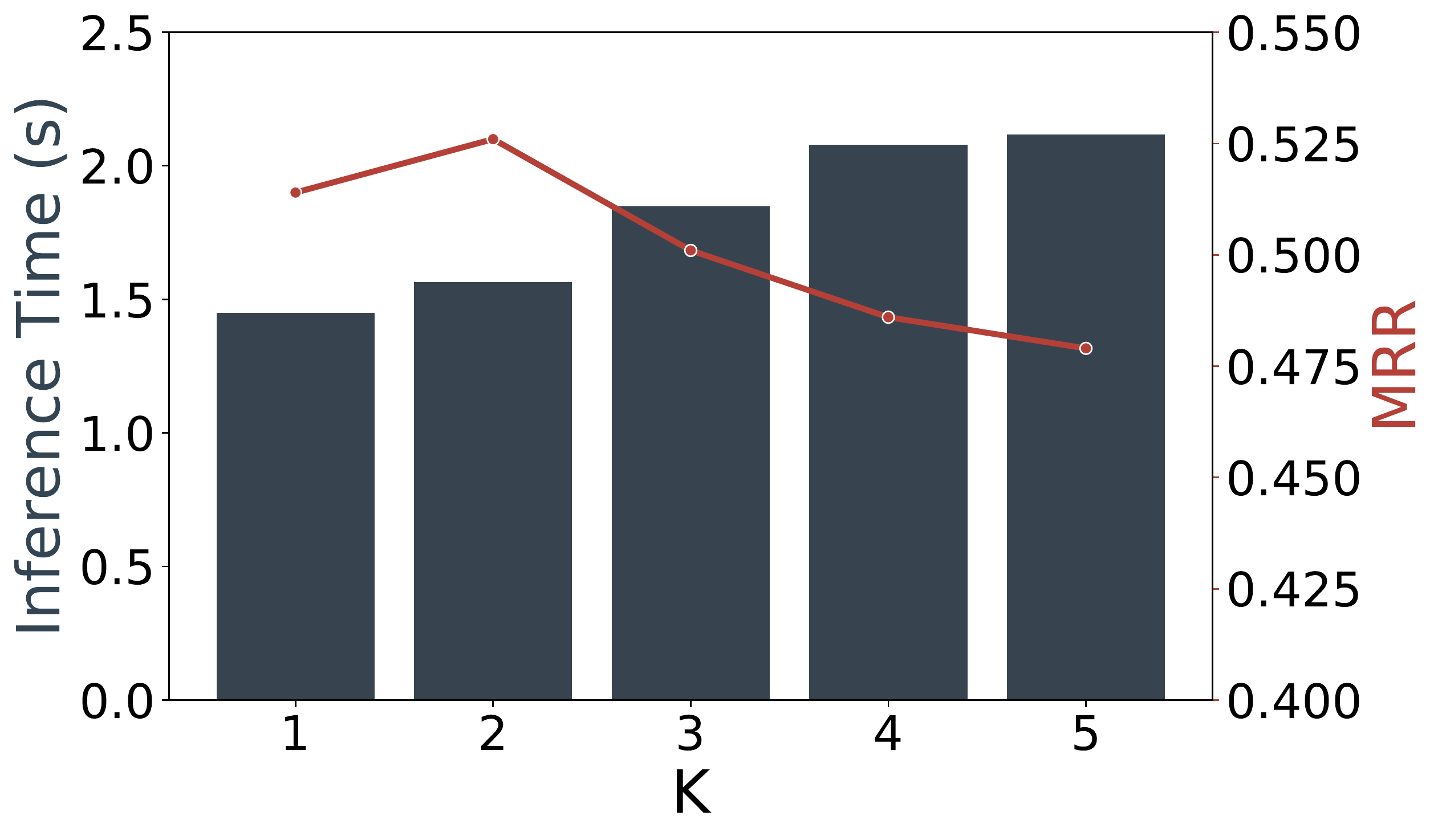}
    \vspace{-0.8em}
    \caption{Effect of $K$ with model $(\mathbb{D}^{100})^4\mathbb{E}^{100}$ on dataset WN18RR.}
    \label{fig:k}
\vspace{-1em}
\end{figure}

We show the effects of the number of active spaces $K$ in Figure \ref{fig:k}. We observe that, compared with $N$, $K$ has a relatively higher impact on the model performance and inference time cost. On WN18RR, a small $K$ (e.g., 1, 2) value can lead to the optimal performance.

\section{Conclusion}

We proposed switch space, a framework for learning expressive product spaces. Switch spaces make use of a sparse gating mechanism to ensure incoming data to be handled by suitable spaces, allowing greater extend of specification in representation. We apply switch space to real world tasks (i.e., knowledge graph completion and item recommendation) and it obtains state-of-the-art performances on benchmark datasets, while being efficient. We believe that switch spaces hold promise for more expressive and generalizable representation learning.

\bibliography{example_paper}
\bibliographystyle{icml2020}




\end{document}